\documentclass[conference]{IEEEtran}
\IEEEoverridecommandlockouts
\usepackage{cite}
\usepackage{amsmath,amssymb,amsfonts}
\usepackage{algorithmic}
\usepackage{graphicx}
\usepackage{textcomp}
\usepackage[table,xcdraw]{xcolor}
\usepackage{url}
\usepackage{multirow}
\usepackage{booktabs}
\usepackage{balance}

\begin{document}

\title{X-Ray Image Compression Using Convolutional Recurrent Neural Networks\\
\thanks
}

\author{Asif Shahriyar Sushmit$^{1}$,
        Shakib Uz Zaman$^{2}$,
        Ahmed Imtiaz Humayun$^{1}$,
        Taufiq Hasan$^{1}$,\\ 
        and Mohammed Imamul Hassan Bhuiyan$^{1,2}$
        \thanks{$^{1}$mHealth Research Group, Department of Biomedical Engineering (BME), Bangladesh University of Engineering and Technology (BUET), Dhaka - 1205, Bangladesh. Email: {\tt\scriptsize taufiq@bme.buet.ac.bd}}%
\thanks{$^{2}$Department of Electrical and Electronics Engineering, Bangladesh University of Engineering and Technology, Dhaka - 1205, Bangladesh Email: {\tt\scriptsize imamul@eee.buet.ac.bd }}}

\maketitle
\begin{abstract}
In the advent of a digital health revolution, vast amounts of clinical data are being generated, stored and processed on a daily basis. This has made the storage and retrieval of large volumes of health-care data, especially, high-resolution medical images, particularly challenging. Effective image compression for medical images thus plays a vital role in today‘s healthcare information system, particularly in teleradiology. In this work, an X-ray image compression method based on a Convolutional Recurrent Neural Networks (\emph{RNN-Conv}) is presented. The proposed architecture can provide variable compression rates during deployment while it requires each network to be trained only once for a specific dimension of X-ray images. The model uses a multi-level pooling scheme that learns contextualized features for effective compression. We perform our image compression experiments on the National Institute of Health (NIH) ChestX-ray8 dataset and compare the performance of the proposed architecture with a state-of-the-art RNN based technique and JPEG 2000. The experimental results depict improved compression performance achieved by the proposed method in terms of Structural Similarity Index (SSIM) and Peak Signal-to-Noise Ratio (PSNR) metrics. To the best of our knowledge, this is the first reported evaluation on using a deep convolutional RNN for medical image compression.
\end{abstract}


\section{Introduction}

In recent years, image compression has been an increasingly active area of research in computer vision for its potential applications in a variety of fields. The aim of image compression is to reduce irrelevance and redundancy of an image for storage or transmission at lower bit rate. In the era of digital health \cite{labrique2018best}, effective compression of medical images have become a critical issue for health-care information systems to deal with the ever increasing size of the digital examination files requiring storage, transmission and processing. In developing countries, teleradiology \cite{bashshur2016empirical} and telemedicine have also gained popularity where the specialist physician or radiologist examines the medical documents or images through digital means while a field doctor, rural medical practitioner (RMP) or a medical technician attends to the patient in remote areas. In such applications, effective compression of high-resolution medical images is vital for fast transmission through the internet or cellular network.

Full-resolution image compression \cite{toderici2017full} is fundamentally a challenging task as the compression artifacts can degrade the image quality and alter the contained information. In case of medical images, nearly loss-less compression/decompression methods are needed as even a slight distortion in the image can lead to errors in diagnosis. Traditional image coding standards (e.g., JPEG and JPEG2000) attempt to distribute the available bits for every non-zero quantized Discrete Cosine Transform (DCT) coefficients in the whole image. As the compression ratio increases, the Bits Per Pixel (BPP) decreases resulting from the use of larger quantization steps, which causes the decoded image to show blocking artifacts \cite{wang2000blind} or noise. A considerable amount of effort has been devoted to decoded image enhancement for deblocking and denoising \cite{wu2002medical,lu1996image}. 

While traditional approaches are generally used for medical image compression \cite{bruylants2015wavelet} \cite{shrestha2010hybrid}, methods incorporating dimensionality reduction have also been applied with convincing results \cite{taur1996medical}. As there are multiple forms of 2-D (histopathological images, X-ray) and 3-D (computed tomography, magnetic resonance imaging) medical image data, classical methods show varying performance across imaging modalities \cite{simek2008gpu,takaya1995information}. 
In recent times, researchers have examined deep learning based image compression for its high performance with respect to minimal loss and higher compression rate. 
Although deep autoencoders are difficult to optimize, deep neural networks with sub-pixel architecture exhibit reasonable results \cite{theis2017lossy,dumas2018autoencoder} in image compression. \cite{theis2017lossy} argues that autoencoder based approaches outperform other classical and experimental compression methods. Recent work has also focused on building a global conceptual compression system using a recurrent variational autoencoder \cite{gregor2016towards}. Though it is convenient for non-medical images, medical image compression requires a more accurate reconstruction due to the diagnostic purposes.

Recurrent Neural Network (RNN) based architectures have also been used for full-resolution image compression \cite{toderici2017full}, which was partially adapted by \cite{kar2018fully} to compress mammogram images of $256\times256$ resolution using a non-recurrent framework. Toderici et al. \cite{toderici2017full} showed that their \emph{RNN-Conv} based methods outperform JPEG compression across a range of bit-rates with superior performance in terms of compression rate, loss and similarity on the Kodak dataset. In this work, we propose a \emph{RNN-Conv} based deep learning architecture that outperforms Toderici et al. \cite{toderici2017full} in a medical image compression task. We draw comparison with the best performing LSTM one-shot framework proposed by \cite{toderici2017full} in terms of similarity and signal to noise ratio.
\begin{figure*}[t]
\includegraphics[width=0.85\linewidth]{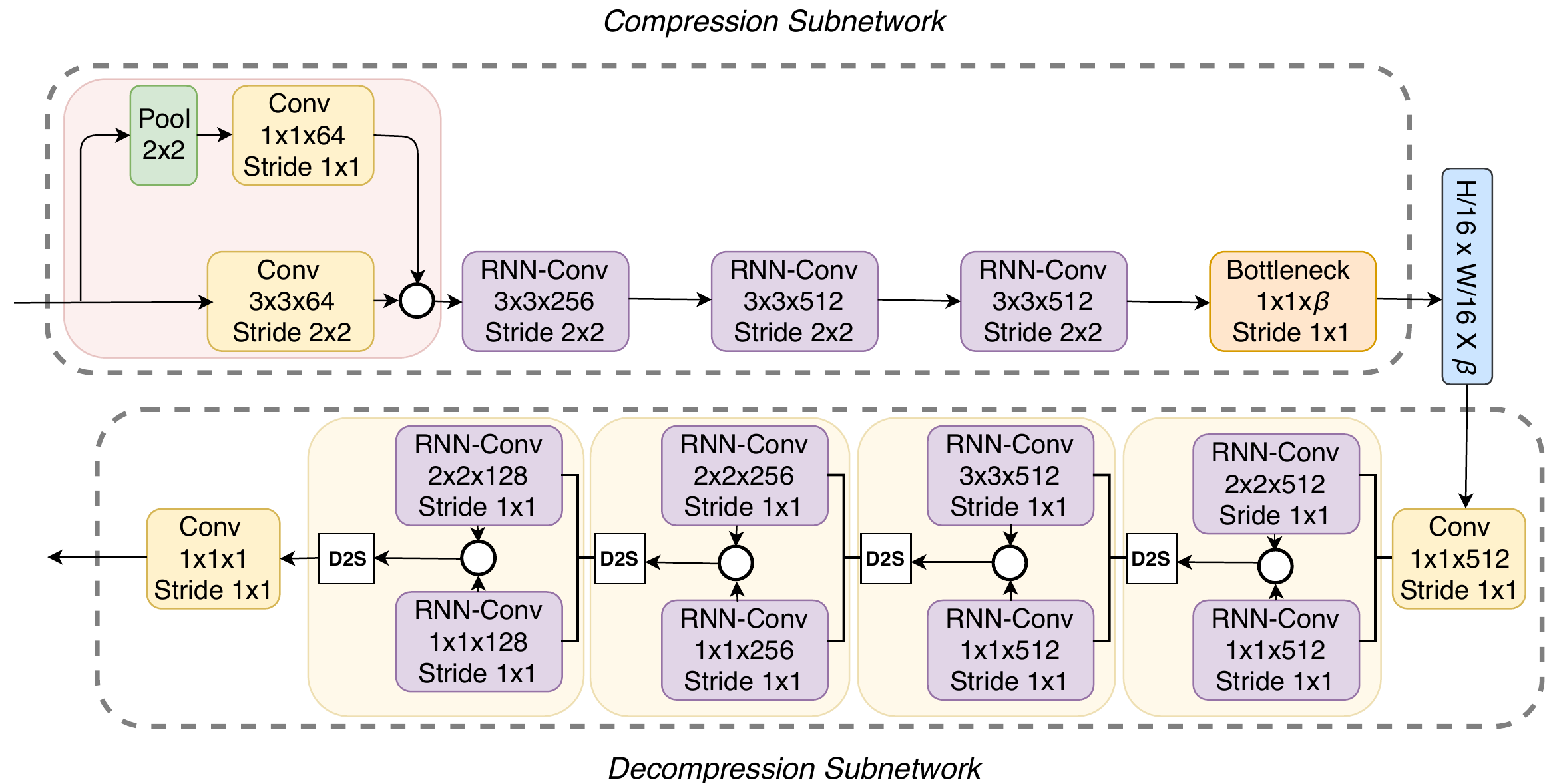}
\centering
\caption{A higher level overview of the proposed Convolutional RNN architecture for X-ray image compression and decompression.}
\label{ProposedModel}
\end{figure*}
\section{X-Ray Image Dataset} \label{dataset}
In our experiments, we use the NIH Clinical Center's Chest X-Ray8 Dataset \cite{wang2017chestx} to evaluate the image compression performance of our model. The dataset is comprised of $112,120$ high quality X-ray images obtained from $30,806$ unique patients and was originally collected for automated lung disease classification tasks. The radiology reports of this dataset are not publicly available, which is not required for our compression task. This dataset contains sufficient data variability and diagnostic complexity, which makes it suitable for performing image compression and reconstruction experiments for the evaluation of the proposed architecture.
\section{Proposed Method} \label{method}
In this section, we describe the proposed \emph{RNN-Conv} network architecture for medical image compression. The network comprises of an encoder $E(\cdot)$ termed compression sub-network, a bottleneck layer $B(\cdot)$ and a decoder $D(\cdot)$ termed decompression sub-network, where both $D(\cdot)$ \& $E(\cdot)$ contain recurrent units. An overview of the proposed compression-decompression architecture is shown in Fig. \ref{ProposedModel}. The input images are first encoded and the bottleneck layer output is stored as a data tensor. The compressed data can subsequently be decompressed using the decoder.
We can compactly represent our network operation as follows:
\begin{equation}
    \mathbf{y}_i = D(B (E (\mathbf{x}_i ) ) )    
\end{equation}
Here, $\mathbf{x}_{i}$ and $\mathbf{y}_{i}$ are matrices of dimension $N\times N$ that represent the original and reconstructed medical image, respectively.

\subsection{Architecture Description}
Convolutional neural networks are inspired by the inter-connectivity of biological neurons in the visual cortex of animals. It is widely deployed in competitive engineering solutions that comapre favoribly with other methods \cite{bottou1994comparison}. On the other hand, Long Short Term Memory (LSTM) units- a variant of RNNs- has been commonly used in time-series modeling since its inception \cite{hochreiter1997long}.
Our network consists of 2D Convolutional Recurrent Neural Networks (\emph{RNN-Conv}) with LSTM units, detailed in \cite{xingjian2015convolutional}.
The front-end of the compression sub-network has two branches; one that performs $2\times2$ strided convolution with a $3\times3$ kernel, while the other performs a $2\times2$ pooling operation prior to convolution with a $1\times1$ kernel. The pooling operation prior to convolution ensures that higher level features encapsulating salient properties of the medical images are learned. The outputs of these two branches are then averaged and passed through $3$ \emph{RNN-Conv} layers with kernel size of $3\times3$ and stride $2\times2$ each. The final CNN layer of the encoder is the bottleneck layer that controls the compression ratio $(r)$ via its number of kernels $\beta$ where, $r = 256/\beta$.

The decompression sub-network takes a tensor of shape $P\times Q \times \beta$ as input. It starts off by performing a linear combination operation along the depth with a Convolutional layer of $512$, $1\times1$ kernels, creating a $P\times Q \times 512$ tensor. The tensor is subsequently passed through $4$ deconvolution modules each with two input branches. Each branch contains an \emph{RNN-Conv} layer with kernel size $2\times2$ and $1\times1$ stride. The output from each branch is averaged, following which a depth-to-space (D2S) operation is performed to spatially upsample the data by a factor of $4$, while reducing the depth (i.e., an input tensor of size $8\times8\times512$ is rearranged to an output of size $16\times16\times128$). The output from the final deconvolution module undergoes another linear combination operation to form the final reconstructed image.



\subsection{Model Training}
The encoder and decoder is trained jointly with a batch-size of $16$. $L1$ loss is optimized using \emph{Adam} optimizer. Both conventional \cite{srivastava2014dropout} and RNN dropout \cite{gal2016theoretically} was trialed which failed to yield significant improvement. Every \emph{RNN-Conv} layer performs padded convolution while the other CNN layers don't. \emph{ReLU} is used as the activation function after each neuron. The output of the bottleneck layer (Fig. \ref{ProposedModel}) is the compressed form of the input medical image. The number of parameters of the resulting model is $62,080,312$.


\begin{table}[h]
\centering
\caption{Performance evaluation on data segments of size $128\times128$}
\label{128data}
\resizebox{\linewidth}{!}{%
\begin{tabular}{c|c|c|c|cc}
\toprule
Image Size    &       Compression        &       Approach      &    SSIM       &   PSNR     \\
 &   Ratio&                 &        &  (dB) \\
\hline  
\multirow{9}{*}{$128\times128$} &   & Toderici \emph{et al.} \cite{toderici2017full} & 0.9387 & 32.1130 \\
            &  2                    & JPEG 2000       & 0.9801 & 25.1276 \\
            &                     & Proposed        & 0.9645 & 36.0152 \\
\cline{2-5}  
           &  & Toderici  \emph{et al.} \cite{toderici2017full} & 0.9119 & 30.8696 \\
            &        4              & JPEG 2000       & 0.9681 & 25.0529 \\
             &                     & Proposed        & 0.9592 & 35.9795 \\
\cline{2-5}
            &   & Toderici  \emph{et al.} \cite{toderici2017full} & 0.8970 & 30.4720  \\
            & 8                     & JPEG 2000       & 0.9409 & 24.8608 \\ 
            &                      & \textbf{Proposed}        & \textbf{0.9579} & \textbf{35.9325} \\ \bottomrule
\end{tabular}%
}
\vspace{3mm}
\centering
\caption{Performance evaluation on data segments of size $256\times256$}
\label{256data}
\resizebox{\linewidth}{!}{%
\begin{tabular}{c|c|c|c|c}
\toprule
Image Size    &       Compression        &       Approach      &    SSIM       &   PSNR      \\
 &   Ratio&                 &        &   (dB)     \\
\hline  
\multirow{9}{*}{$256\times256$}     &   & Toderici \emph{et al.} \cite{toderici2017full} & 0.9013 & 31.9197 \\
            &   2                   & JPEG 2000       & 0.9783 & 25.1051 \\
             &                     & Proposed        & 0.9525 & 34.9721 \\
\cline{2-5}
            &  & Toderici \emph{et al.} \cite{toderici2017full} & 0.8933 & 30.2135 \\
            &       4               & JPEG 2000       & 0.9713 & 25.0797 \\
            &                      & Proposed        & 0.9517 & 34.9002 \\
\cline{2-5} 
            &   & Toderici \emph{et al.} \cite{toderici2017full} & 0.8755 & 30.015   \\
            &   8                   & JPEG 2000       & 0.9500 & 25.0208 \\ 
            &                     & \textbf{Proposed}        & \textbf{0.9509} & \textbf{34.8701} \\ \bottomrule
\end{tabular}%
}
\end{table}



\section{Performance Evaluation}
\subsection{Experimental Setup}
The original images of the NIH dataset are of resolution $1024\times1024$. Due to computational constrains, we could not perform experiments using full resolution images directly. Instead, we trained our network on $N\times N$ sized segments of the data, where $N \in \{128, 256\}$. 
For each value of $N$, we train and test the model's performance using randomly sampled $N\times N$ sized cropped images from the original dataset. 
We obtain $20,000$ samples for training and $1,000$ samples for validation, from each segment size. We also prepare a test set of $500$ samples for model evaluation. To ensure that our algorithm learns patient invariant features, the training, validation and test set image segments are obtained from mutually exclusive patients.  

\subsection{Experimental Results} \label{resDisc}
We contrast the performance of our model with an implementation of \cite{toderici2017full}, where the authors use successive 2D LSTM convolutions to perform compression, and converted the information utilizing a binarizer. The decompression sub-network of \cite{toderici2017full} also employed 2D LSTM convolutions. 
We note that, there is no consensus in the image compression literature on which performance metric best correlates with human perception.
In our experiments, we select the Structural Similarity Index Metric (SSIM) and Peak Signal to Noise Ratio (PSNR) as evaluation metrics. SSIM provides a similarity score between $0$ and $1$ and PSNR is a measure of noise-level quantified in decibels ($dB$). In both cases, higher values imply a closer match between the test and reference images.

We can observe in Tables \ref{128data} and \ref{256data} that our model outperforms the implementation of Toderici et al \cite{toderici2017full} and JPEG 2000 for a compression ratio of 8 on both $128\times128$ and $256\times256$ test sets. For the higher resolution $256\times256$ image segments, we see a $4.85$ $dB$ increase in PSNR and a $8.61\%$ relative increase in SSIM compared to the one-shot LSTM Conv model of Toderici et al. \cite{toderici2017full}. 



\begin{figure*}[t]
\includegraphics[width=.75\linewidth]{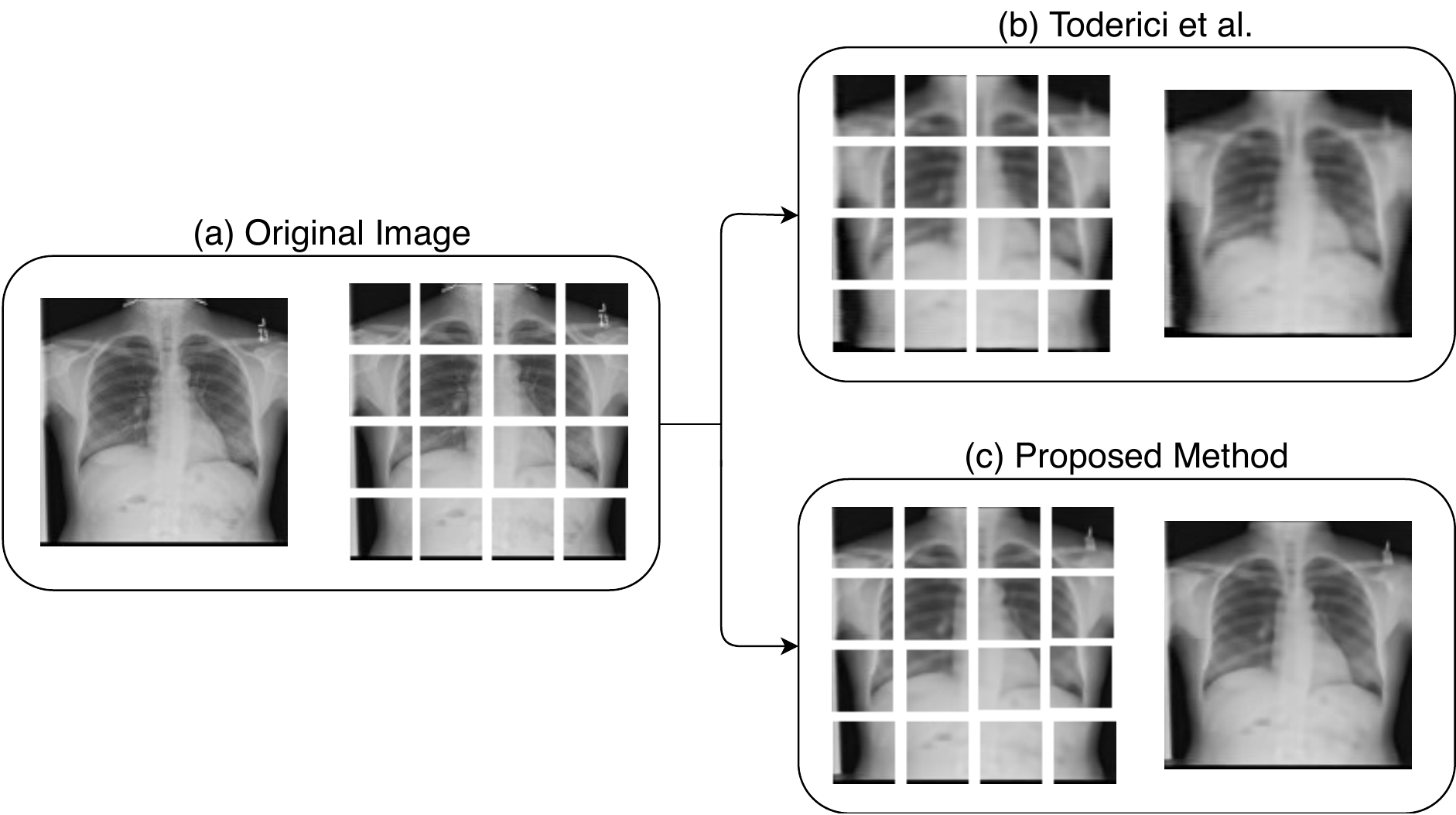}
\centering
\caption{The original and compressed X-ray images obtained using the proposed network and \cite{toderici2017full}. The full-size images are cropped into $16$ equally sized mutually exclusive samples for processing. (a) The original full size image and cropped version. (b) Compressed and decompressed image patches and subsequent full-size image obtained using the methods of Toderici et al. \cite{toderici2017full}. (c) Compressed and decompressed image patches and subsequent full-size image obtained using the proposed convolutional RNN. Proposed method shows a sharper X-ray image compared to Toderici et al. \cite{toderici2017full}.}
\label{recon}
\end{figure*}

\subsection{Discussion}
As observed from the results, the performance of our method is promising for low-resolution images. But we can observe the trend of increasing performance for higher compression rates. When comparing the performance of our model with JPEG2000, we can see that for lower compression ratio, despite having a better PSNR, our model has an SSIM score that is below that of JPEG-2000. But for a compression ratio of $8$, our model has a better score in terms of both the performance metrics. Also, perceptually, our model yields reconstructed images that are almost indistinguishable from original ones. We have also experimented with higher compression ratios than $8$, in which cases our model exhibits a steady PSNR value where as the PSNR for JPEG-2000 starts dropping. From Table \ref{128data} and \ref{256data} we can also see a drop in performance with an increase in image dimensions. While JPEG-2000 outperforms our method for lower compression ratios, the difference diminishes as compression ratio is increased.

As the patches are of low-resolution, it is difficult to compare the difference of two reconstructions visually. Therefore, to demonstrate the performance of the proposed model compared with \cite{toderici2017full},  we show the images generated after successive compression and decompression using the two methods in Fig. \ref{recon}. In total, the performance of 6 different variants of the network architecture is investigated, while the best performing model is proposed in this paper. 
We can further compress the encoded tensors by applying loss-less source coding techniques such as \emph{GZip}. Through this, we are able to further compress the encoded images by a factor of $4$.


\section{Conclusion}
In this work, we have presented an Convolutional RNN based architecture for medical image compression, specifically for chest X-rays. The proposed method provides superior performance compared to both traditional and a state-of-the-art methods. The reconstructed images were shown to be visually indistinguishable from original images and contains very low-levels of noise that is vital for high quality medical image storage and transmission. 
The model has provided a consistent level of performance in terms of PSNR for high compression ratios, which are comparable to lower compression rates.
Our proposed model exhibits a $4.85$ $dB$ increase in PSNR and a $8.61\%$ relative increase in SSIM compared to the one-shot LSTM Conv model of Toderici et al. \cite{toderici2017full} for a compression ratio of $8$. We also observe a $9.85$ $dB$ increase in PSNR compared to standard JPEG-2000 compression.


\section*{Acknowledgment}
We would like to thank the Department of EEE \& BME, BUET and Brain Station 23 (Dhaka, Bangladesh) for supporting this research. The Titan Xp GPU used for this work was donated by the NVIDIA Corporation.
\bibliographystyle{IEEEtran}
\bibliography{refs.bib}

\end{document}